\documentclass[10pt,twocolumn,letterpaper]{article}

\usepackage[pagenumbers]{cvpr} 

\definecolor{cvprblue}{rgb}{0.21,0.49,0.74}
\usepackage[pagebackref,breaklinks,colorlinks,citecolor=cvprblue]{hyperref}
\usepackage{algorithmic}
\usepackage[ruled]{algorithm2e}
\usepackage{multirow}

\title{Random Sampling for Diffusion-based Adversarial Purification}

\author{%
	Jiancheng Zhang $^{1}$, Peiran Dong $^{2}$, Yongyong Chen $^{3}$, Yin-Ping Zhao $^{1}$, Song Guo $^{4}$~\\
	$^{1}$ Northwestern Polytechnical University,  $^2$ The Hong Kong Polytechnic University, \\ $^3$ Harbin Institute of Technology (Shenzhen), $^4$ The Hong Kong University of Science and Technology
}

\begin{document}
\maketitle
\begin{abstract}
Denoising Diffusion Probabilistic Models (DDPMs) have gained great attention in adversarial purification. Current diffusion-based works focus on designing effective condition-guided mechanisms while ignoring a fundamental problem, i.e., the original DDPM sampling is intended for stable generation, which may not be the optimal solution for adversarial purification. Inspired by the stability of the Denoising Diffusion Implicit Model (DDIM), we propose an opposite sampling scheme called random sampling. In brief, random sampling will sample from a random noisy space during each diffusion process, while DDPM and DDIM sampling will continuously sample from the adjacent or original noisy space. Thus, random sampling obtains more randomness and achieves stronger robustness against adversarial attacks.
Correspondingly, we also introduce a novel mediator conditional guidance to guarantee the consistency of the prediction under the purified image and clean image input. To expand awareness of guided diffusion purification, we conduct a detailed evaluation with different sampling methods and our random sampling achieves an impressive improvement in multiple settings. Leveraging mediator-guided random sampling, we also establish a baseline method named DiffAP, which significantly outperforms state-of-the-art (SOTA) approaches in performance and defensive stability. Remarkably, under strong attack, our DiffAP even achieves a more than $20\%$ robustness advantage with $10\times$ sampling acceleration. 
\url{https://github.com/ZhangJC-2k/RandomSampling}

\end{abstract}    
\section{Introduction}
\label{sec:intro}
Advances in deep neural networks (DNNs) \cite{AlexNet,resnet,Swin} have led to some practical applications \cite{face_recognition,text-to-image_diff,gpt4,differential_privacy,cyber_security} but also to some security concerns \cite{security_deepL}. The results of adversarial attack methods \cite{autoattack, attack1, BPDA} show that DNNs are very sensitive to some artificial perturbations and can easily be guided by these imperceptible perturbations to generate inappropriate results. In the face of potential threats, both adversarial training \cite{adv_training1, adv_training2} and adaptive test-time \cite{adaptive_testtime1, adaptive_testtime2} defense approaches have been specifically designed. The former leverages training sets containing adversarial samples to improve the robustness of DNNs during training, which comes at an additional cost and only defends against training-type attacks. The latter adaptively mitigates the effects of perturbations at test time, which is more efficient and flexible. 

Among the adaptive test-time defenses, adversarial purification via generative models \cite{Defense_GAN, DiffPure, ap_score_based, ap_Energy-Based_Models} utilizes the data distribution prior \cite{score_based_sde} in the generated model to remove adversarial perturbations, which attracts wide attention. As the pioneering generative model, Denoising Diffusion Probabilistic Models (DDPMs) \cite{ddpm, score_based_sde, ddim} has been explored for adversarial purification \cite{DiffPure, GDMP, MimicDiffusion} and achieve promising effect. Diffusion models perturb the data into the noise domain and learn the gradient of data distributions during the forward training process. Then the generated data is iteratively sampled from the pure Gaussian noise in reverse inference processes. To employ it in purification, the forward noising process and reverse sampling process are mimicked in an adversarial image. However, the original DDPM sampling is designed for stable generation, which may not be the optimal solution for adversarial purification.

With a theoretical guarantee, DiffPure \cite{DiffPure} preliminarily demonstrated the potential of the diffusion model as a purifier for adversarial image purification. 
To unlock the potential of diffusion-based purification, conditional guidance \cite{GDMP, MimicDiffusion} makes it possible to purify from pure Gaussian noise, which further eliminates the influence of the adversarial attack. But a new challenge also arises: How to guarantee the consistency of the prediction under the purified image and clean image input? In addition, for unconditional diffusion purification methods, Robustness Evaluation \cite{RobustEvaluation} showed that while robustness steadily improved as the number of forward steps increased, the consistency of predictions continued to decline. However, there is a lack of comprehensive evaluation and configuration recommendations for condition-guided diffusion methods.

Inspired by the stability of the Diffusion Denoising Implicit Model (DDIM) \cite{ddim}, we have developed an opposite sampling method, i.e., Random Sampling for adversarial purification tasks. This method aims to enhance randomness in the diffusion sampling process to establish a specialized sampling method for adversarial purification. Briefly, random sampling will sample from a random noisy space during each diffusion process, while DDPM and DDIM sampling will continuously sample from the adjacent or original noisy space. Thus, random sampling obtains more randomness and achieves stronger robustness against adversarial attacks. To show this, we also present a conceptual illustration of the sampling trajectory. As shown in Fig.~\ref{fig:sampling_trajectory} (a) and (b), the sampling process of DDIM and DDPM is closer to the forward process and the sampling points are adjacent, which is easily predicted by attack methods. By contrast, the sampling trajectory of random sampling is completely random and unpredictable, as plotted in Fig.~\ref{fig:sampling_trajectory} (c).

When random sampling stands as a tailored enhancement technique to improve the randomness of diffusion-based adversarial purification, matching condition guidance is necessary to ensure that the prediction does not deviate too much from the input. Thus, we propose a mediator guidance to achieve effective condition guidance, which also can optimize the prediction consistency between clean image and purified image. In a nutshell, gradient-based guidance is employed to implement accurate conditional guidance in the mediator variable rather than a noisy sampling point. The mediator guidance can avoid gradient bias and significantly improve the consistency of the purification model, which is even close to inputting clean images in the classifier. Meanwhile, it also greatly enhances the stability of diffusion sampling, making the purification method perform well in different Settings without adjustment. 

Based on the mediator guidance, we conduct a detailed evaluation with different sampling methods to expand awareness of condition-guided diffusion purification. In evaluation, our random sampling achieves an impressive improvement with commonly used DDPM sampling and we also find a challenging asynchronous attack. Leveraging mediator-guided random sampling, we also establish a baseline method named DiffAP, which significantly outperforms SOTA approaches in performance and defensive stability with fewer steps. Remarkably, under strong attack, our DiffAP even achieves a more than $20\%$ robustness advantage with $10\times$ sampling acceleration.
Overall, the contributions of our work are summarized as follows:
\begin{itemize}

\item We present a versatile random sampling, which enhances the randomness of the sampling process and is a robust solution for diffusion-based adversarial purification.

\item To maintain consistency between clean and purified sample predictions, we propose mediator guidance, which also improves the stability of guided methods.

\item An evaluation of guided diffusion purification is conducted and provides a challenging attack scheme.

\item We introduce a baseline method DiffAP, which can significantly outperform SOTA approaches with $10\times$ sampling acceleration.

\end{itemize}

\vspace{-2mm}
\section{Background}
\label{background}
\vspace{-1mm}
\subsection{Diffusion-based Adversarial Purification}
\vspace{-2mm}
Adversarial purification methods \cite{Robustifying_models, PixelDefend,ap_Energy-Based_Models,ap_gaussian_mixture_vae,ap_score_based, Defense_GAN} aims to introduce additional prior information to process potentially malicious perturbations in the input. Among them, diffusion-based methods \cite{DiffPure, GDMP, MimicDiffusion, Robustbench} utilize the learned data distribution prior \cite{score_based_sde} of pretrained diffusion models  \cite{ddpm} to purify the adversarial examples. 

DiffPure \cite{DiffPure} preliminarily diffuses the adversarial input with a small amount of noise following a forward diffusion process, and then recovers the clean image through a reverse generative process. Robustness Evaluation \cite{RobustEvaluation} focused on the performance of unconditional diffusion purification \cite{DiffPure} and gave some recommendations for defense and attack settings. GDMP \cite{GDMP} submerges the adversarial perturbations with gradually added Gaussian noises and then utilizes the conditional guidance to restore the purified image, which achieves more better purification effect. MimicDiffusion \cite{MimicDiffusion} mimics the trajectory of the diffusion model with clean inputs to reduce the effect of adversarial perturbations and proposes two guidance based on Manhattan distance. Naturally, effective conditional guidance has become the focus of the current research. However, there is a lack of basic evaluation and configuration recommendations for conditional diffusion methods.

\subsection{Denoising Diffusion Probabilistic Models}
\vspace{-2mm}
Diffusion model \cite{ddpm} includes forward noising process and reverse sampling process. The forward process gradually added Gaussian noises to the clean image ($\mathbf{x}_0$) and eventually got a noisy image or Gaussian noise ($\mathbf{x}_t$), which can be formulated as:
\begin{equation}
    \mathbf{x}_t = \sqrt{\Bar{\alpha}_t}\mathbf{x}_0 + \sqrt{1-\Bar{\alpha}_t}\epsilon,
    \label{eq1}
\end{equation}
where $\alpha_t = 1 - \beta_t$, $\Bar{\alpha}_t = \prod_{k=1}^{t} \alpha_k$, $t$ means the current time steps, $\epsilon \sim \mathcal{N}(0,\mathbf{I})$ and ${\beta_t}$ is a continuously increasing sequence. Given the number of forward steps $T$, the reverse process iteratively sample $\mathbf{x_{t-1}}$ from $\mathbf{x}_T$, which can be formulated as:
\begin{equation}
    \mathbf{x}_{t-1} = \frac{1}{\sqrt{\alpha_t}}(\mathbf{x}_t - \frac{\beta_t}{\sqrt{1-\Bar{\alpha}_t}}\epsilon_{\theta}(\mathbf{x}_t,t))+\sqrt{\beta_t}\epsilon_t,
    \label{eq2}
\end{equation}
where $\epsilon_{\theta}(\mathbf{x}_t,t)$ is the predicted noise of DDPMs at $t$ step and $\epsilon_t \sim \mathcal{N}(0,\mathbf{I})$ is standard Gaussian noise. Ultimately, we can generate the clean image $\mathbf{x}_0$ from the initial sampling point $\mathbf{x}_T$, which also is directly utilized in previous diffusion-based purification methods \cite{DiffPure,Robustbench,GDMP,MimicDiffusion}. 
Considering that stability and speed are the core requirements for the generation task, the sampling method of Eq.~\eqref{eq2} may not be the optimal solution for adversarial purification when the defenses benefit from randomness. Thus, a fundamental question arises: is there a more robust sampling method for adversarial purification?

\section{Methods}
\label{sec:methods}
Inspired by the stability of the Denoising Diffusion Implicit Model (DDIM) \cite{ddim}, we revisit the DDIM sampling process and propose
an opposite sampling scheme called random sampling, which may be a robust solution.

\subsection{Revisit Denoising Diffusion Implicit Model}
\noindent\textbf{Joint Distribution Family}.
A key observation of DDIM is that the DDPM objective in the form of $L_{\gamma}$ only depends on the marginals $q(x_t|x_0)$,  but not directly on the joint $q(x_{1:T}|x_0)$. Since there are many inference distributions (joints) with the same marginals, they explore alternative inference processes that are non-Markovian, and present the joint distribution family as follows:
 \begin{equation}
 \begin{aligned}
\mathbf{x}_{t-1}=&\sqrt{\bar{\alpha}_{t-1}}\left(\frac{\mathbf{x}_{t}-\sqrt{1-\bar{\alpha}_{t}} \mathbf{\epsilon}_\theta(\mathbf{x}_{t}, t)}{\sqrt{\bar{\alpha}_{t}}}\right)\\
+&\sqrt{1-\bar{\alpha}_{t-1}-{\sigma^2_{t}}} \cdot \mathbf{\epsilon}_\theta(\mathbf{x}_{t}, t)+\sigma_{t} \mathbf{\epsilon}_{t},
\label{eq7}
\end{aligned}
 \end{equation}
where the term inside the first bracket can be treated as denoised image $\Tilde{\mathbf{x}}_0$ predicted via current $\mathbf{x}_t$ and the magnitude of $\sigma_{t}$ controls the variance of new random noise.

\noindent\textbf{Difference between DDPM and DDIM}.
When $\sigma_t = \sqrt{(1 - \alpha_{t-1}) / (1 - \alpha_t)} \sqrt{1 - \alpha_t / \alpha_{t-1}}$ for all $t$, Eq.~\eqref{eq7} can be rewritten as Eq.~\eqref{eq2}, and the generative process becomes a DDPM. When $\sigma \to 0$, we reach an extreme case where as long as we observe $x_0$ and $x_t$ for some $t$, then $x_{t-1}$ becomes known and fixed. The resulting model becomes an implicit probabilistic model \cite{implicit_generative}, where samples are generated from latent variables with a fixed procedure (from $x_T$ to $x_0$), which is also DDIM \cite{ddim}. Therefore, the core difference between the two is that DDPM will introduce some new noise into the original noisy space during each sampling while DDIM is fixed in the original noisy space sampling. Naturally, the former has stronger randomness and is therefore widely used while the latter is ignored by previous methods \cite{DiffPure,GDMP,MimicDiffusion}, which is also consistent with the results in the Robustness Evaluation \cite{RobustEvaluation}. 

\subsection{Random Sampling}
Inspired by the observation, we try to maximize the randomness in the sampling process to propose a new sampling scheme. To make it easier to understand and test, we replace the value $\sigma_{t}$ with a proportion $k_t \in [0,1]$ and rewrite the Eq.~\eqref{eq7} in the following form:
 \begin{equation}
 \begin{aligned}
\mathbf{x}_{t-1}=&\sqrt{\bar{\alpha}_{t-1}}\left(\frac{\mathbf{x}_{t}-\sqrt{1-\bar{\alpha}_{t}} \mathbf{\epsilon}_\theta(\mathbf{x}_{t}, t)}{\sqrt{\bar{\alpha}_{t}}}\right)\\
+&\sqrt{1-k_t} \sqrt{1-\bar{\alpha}_{t-1}} \cdot \mathbf{\epsilon}_\theta(\mathbf{x}_{t}, t)+\sqrt{k_t}\sqrt{1-\bar{\alpha}_{t-1}} \mathbf{\epsilon}_{t}.
\label{eq4}
\end{aligned}
 \end{equation}
\noindent\textbf{Random Noise Rate}.
We explore the effect of new noise on robustness by varying the rate of new noise $k_t$ from 0 to 1, resulting in the changes of its total variance ranging from 0 to $\sqrt{1-\bar{\alpha}_{t-1}}$. 200 forward steps are used for both attack and defense, and we set five denoising steps for attack and defense for all experiments.

\begin{figure}[!t]
    \begin{center}
        \includegraphics[width=1.05\linewidth]{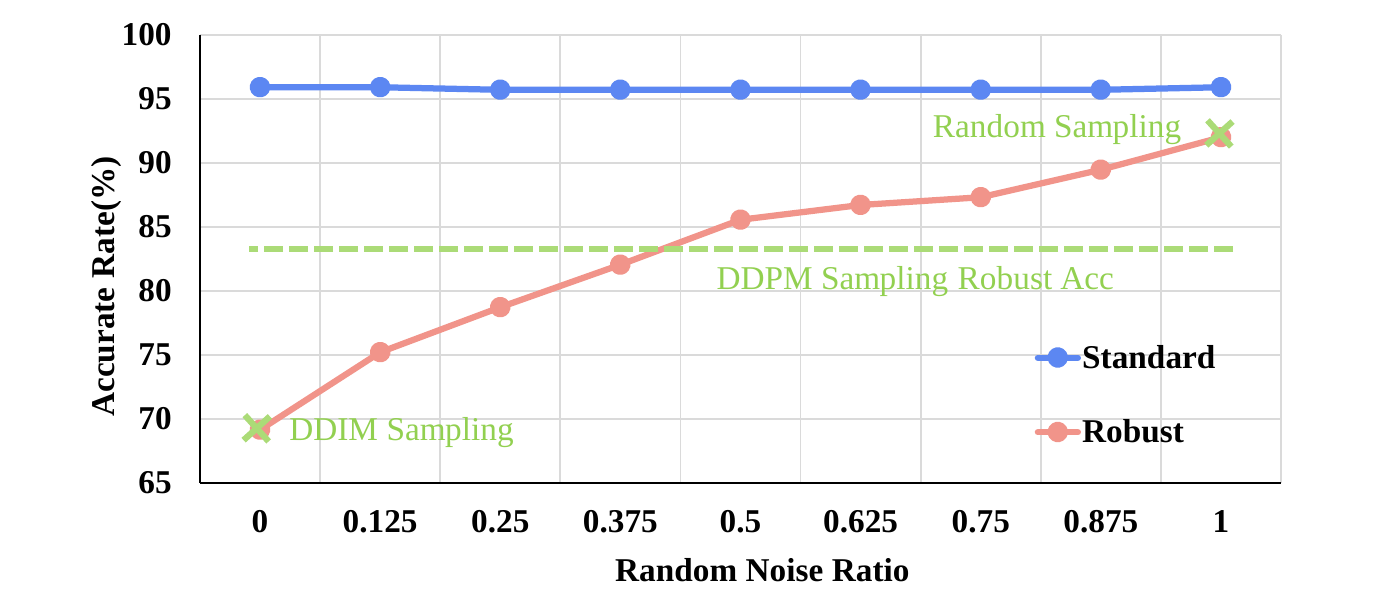}
        \vspace{-8mm}
        \caption{Standard and robust accuracy as we change the rate of random noise against PGD+EOT $\ell_\infty (\epsilon = 8/255)$ on CIFAR-10. Notably, our random sampling achieves a significant advantage over DDPM and DDIM sampling, making it a more robust solution for diffusion-based adversarial purification.}
        \label{fig:noise_rate}
    \end{center}
 \vspace{-10mm}
\end{figure}

As shown in Fig.~\ref{fig:noise_rate}, the robust accuracy continuously increases as the rate of new noise increases since more new noise induces more randomness.
Locations with a random noise ratio of 0 represent the result of DDIM sampling, which is the most stable sampling method and also has the lowest robust accuracy. DDPM sampling has a time-varying noise ratio, so we only report results for its robustness, roughly in the middle. Finally, the locations with a random noise ratio of 1 represent the new sampling scheme, which achieves the strongest randomness and also has the highest robust accuracy. The result intuitively illustrates that The randomness of the sampling process is almost proportional to the robustness of the sampling method.

\noindent\textbf{Sampling Scheme}. Thus, we set $\sigma=\sqrt{1-\bar{\alpha}_{t}}$ to establish our random sampling, which is equivalent to sampling from a new noisy space during each diffusion process and is not affected by the previous sample position.
Thus, our proposed random sampling can be written as follows:
 \begin{equation}
 \begin{aligned}
\mathbf{x}_{t-1}=&\sqrt{\bar{\alpha}_{t-1}}\left(\frac{\mathbf{x}_{t}-\sqrt{1-\bar{\alpha}_{t}} \mathbf{\epsilon}_\theta(\mathbf{x}_{t}, t)}{\sqrt{\bar{\alpha}_{t}}}\right)+\sqrt{1-\bar{\alpha}_{t-1}} \cdot \mathbf{\epsilon}_{t}.
\label{eq5}
\end{aligned}
 \end{equation}

 \begin{figure*}[!t]
    \begin{center}
        \includegraphics[width=1\linewidth]{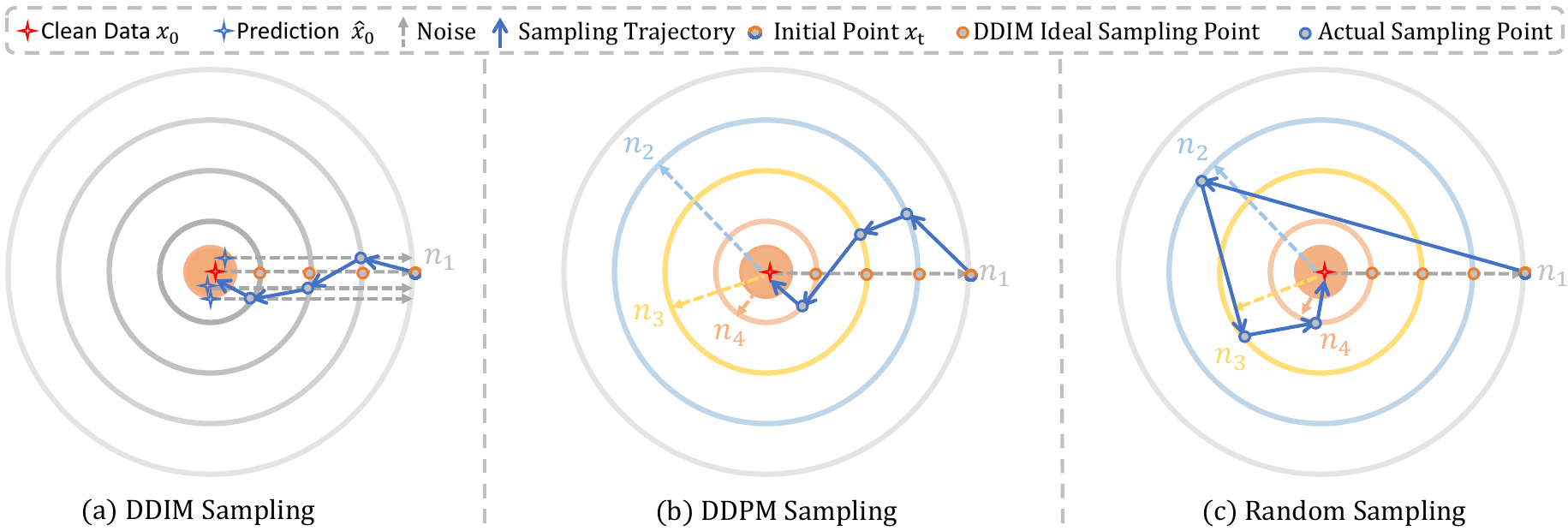}
        \vspace{-5mm}
        \caption{Conceptual illustration of the sampling trajectory of three different diffusion processes. 
        The orange center area represents the distribution of clean samples, and the outer circle represents the sampling space after the noise disturbance. The noise intensity on the same circle is the same when the noise disturbance on the outer circle is greater.
        }
        \label{fig:sampling_trajectory}
    \end{center}
 \vspace{-8mm}
\end{figure*}
 
\noindent\textbf{Sampling Trajectory Analysis}. As shown in Fig.~\ref{fig:sampling_trajectory}, we present a conceptual illustration of the sampled trajectory of different sampling methods. Regardless of the form of the forward process, the final noisy image can be equivalent to adding a noise $n_1$ to the initial point of the reverse sampling process. The ideal DDIM sample is the most stable, but because the estimates of the diffusion model are always biased, the actual DDIM sample will fluctuate. The ideal DDIM when given $x_0$ and $x_t$ the intermediate sampling process should be deterministic, as shown by the orange sampling point in Fig.~\ref{fig:sampling_trajectory}(a). However, because the trained diffusion model is not completely accurate, the actual DDIM sampling will fluctuate, as shown by the blue sampling points in Fig.~\ref{fig:sampling_trajectory}(a). For DDPM sampling, the sampling trajectory will be affected by the addition of new noise, showing greater fluctuations in Fig.~\ref{fig:sampling_trajectory}(b). However because each new sampling point still depends on the previous sampling point, it is still relatively stable. Different from the two, the proposed random sampling only depends on the new sampling noise and is not affected by the previous sampling point, as shown in Fig.~\ref{fig:sampling_trajectory}(c). 
When the trajectory of DDIM and DDPM sampling is distributed around a fixed sampling trajectory (DDIM ideal sampling points), our random sampling method is unconstrained and presents the most randomness in the sampling space, which brings more gain for diffusion-based adversarial purification.

\subsection{Mediator Guidance}
Although the results of Robustness Evaluation \cite{RobustEvaluation} show that condition guidance reduces the robustness of diffusion models, this is a misunderstanding due to the difference in the performance distribution of conditional and unconditional guidance models, which we show in the Sec.~\ref{sec: forwad_step}.

\noindent\textbf{Prediction Consistency}. Conditional guidance is still necessary for diffusion-based adversarial. On the one hand, the more noise added, the more effective the removal of antagonistic perturbations can be, but this also makes the diffusion process difficult to predict. Therefore, the process requires conditional guidance to ensure that the purified image does not deviate too much from the input. On the other hand, although we do not know whether the input is a clean image or an adversarial image, we expect that the accuracy of the clean image input into the prediction model will not be reduced after purification. 
In addition, there are some anomalies in the previously commonly used guidance conditions \cite{MimicDiffusion}, which we will show later and the corresponding theoretical analysis will be added to the appendix. Here, we present a different conditional guidance design from the previous method \cite{GDMP,MimicDiffusion}, which will better guide the diffusion-based purification process. 

\noindent\textbf{Mediator Variable Guidance}.
For the case of DDPM sampling, $p(x_0|x_t)$ has the unique posterior mean \cite{DPS} at
\begin{align}
\label{eq9}
 E[x_0|x_t] &= \frac{1}{\sqrt{{\bar\alpha(t)}}}(x_t - \sqrt{1 - {\bar\alpha(t)}}\nabla_{x_t} \log p_t(x_t))
\end{align}
which happens to be the mediator variable in the $x_t$ update process of Eq.~\eqref{eq5} when $\mathbf{\epsilon}_\theta(\mathbf{x}_{t}, t)$ estimates the gradient of data distribution $\nabla_{x_t} \log p_t(x_t)$ \cite{score_based_sde,ddpm}. Considering that we want this expectation distribution to be as close to the input as possible during each update, we apply gradient-based guidance \cite{GDMP} to the mediator variable as follows:
\begin{equation}
\begin{aligned}
\label{eq:grad_log_bayes}
    \Tilde{\mathbf{x}}_{0,t} \gets \Tilde{\mathbf{x}}_{0,t} - \nabla_{\Tilde{\mathbf{x}}_{0,t}} \log p(x^{adv/clean}|\Tilde{\mathbf{x}}_{0,t})
\end{aligned}
\end{equation}
where $\Tilde{\mathbf{x}}_{0,t}$ is the mediator variable in the $t$ time and $\nabla_{\Tilde{\mathbf{x}}_{0,t}} \log p(x^{adv/clean}|\Tilde{\mathbf{x}}_{0,t})$ could be obtained as follows:
\begin{equation}
\begin{aligned}
    \nabla_{\Tilde{\mathbf{x}}_{0,t}} \log p(x^{adv/clean}|\Tilde{\mathbf{x}}_{0,t};t) = -R_{t}\nabla_{\Tilde{\mathbf{x}}_{0,t}}d(\Tilde{\mathbf{x}}_{0,t},x^{adv/clean}), \\
    \Tilde{\mathbf{x}}_{0,t} = \frac{x_{t}-\sqrt{1-\bar\alpha_t} \mathbf{\epsilon}_\theta(\mathbf{x}_{t}, t)}{\sqrt{\bar\alpha_t}},
    \label{eq11}
\end{aligned}
\end{equation}
where $\mathbf{\epsilon}_\theta(\mathbf{x}_{t}, t)$ is the score estimation~\cite{score_based_sde} through the pre-trained diffusion model with the parameter $\theta$ for $x_{t}$ in the $t$ time, $R_{t}$ is the guided factor related to the $t$ time, and $d(\ast,\ast)$ is the $MSE$, $\ell_1$ norm, $\ell_2$ norm, or other distance metrics.

\noindent\textbf{Selected Conditional Guidance}.
As shown in Fig.~\ref{fig:forward_step2}, a curious defensive failure occurs with the DDIM sampling method, which is even worse than the performance of the unconditionally guided diffusion model. Considering that DDIM sampling is the most stable sampling method, the accurate guided condition may make it more vulnerable to attack than other sampling methods. In addition, previous works \cite{MimicDiffusion, FreeDoM} show that less guidance can reduce the time cost without affecting the performance. Thus, we select the partial phase of the reverse process to add conditional guidance when part of the unconditional diffusion process may mitigate the collapse. Concretely, the whole generation time step is in $[T_{s}(M), T_{s}(M-1),\ldots,T_{s}(0)]$, we implement the guided method in the subsequence $[T_{s}(M~mod~k), T_{s}((M~mod~k)-1),\ldots,T_{s}(0)]$ and vice versa not implementing the guided method. In this way, we avoid implementing the guided method in each step to get away with being specifically attacked, saving some unnecessary gradient calculation, and thus reducing the time cost at the same time.

Algorithm~\ref{algorithm1} summarizes the proposed Mediator-guided Random Sampling. In our method, the hyperparameters include the $R_{T_s(i)}$ and the modulus $k$. For the guided factor, it could be calculated directly without additional constraints. We experimentally find that a fixed constant is enough to achieve stable and effective guided effects, similar to the scheme of \cite{MimicDiffusion}. Ultimately, we select the guided subsequence by setting $k$ as 2, which makes it possible to work with very few diffusion steps such as 10. 

\begin{algorithm}[t]
    \small
    \caption{\label{algorithm1}Mediator-guided Random Sampling}
    \DontPrintSemicolon
    \SetAlgoLined
    \textbf{Require:} Forward Step $T$, Denosing Step $M$, $x^{adv/clean}$ \;
     $T_s=[M*(\frac{T}{M}),(M-1)*(\frac{T}{M}),...,1*\frac{T}{M}, 0]$\;
     $x_T \sim\mathcal{N}(\sqrt{\Bar{\alpha}_T} x^{adv/clean}, (1-\Bar{\alpha}_T) I$) \;
    \For{$i=M$ {\bfseries to} $1$}{
         $\hat{s} \gets \mathbf{\epsilon}_\theta(x_{T_s(i)}, {T_s(i)})$ \;
         $\Tilde{\mathbf{x}}_{0,t} \gets \frac{1}{\sqrt{\bar\alpha_{T_s(i)}}}(x_{T_s(i)} - \sqrt{1 - \bar\alpha_{T_s(i)}}\hat s)$ \;
        \eIf{i mod k = 0}{
         $\Tilde{\mathbf{x}}_{0,t} \gets \Tilde{\mathbf{x}}_{0,t} -  {R_{T_s(i)}}\nabla_{\Tilde{\mathbf{x}}_{0,t}}D(x^{adv/clean}, \Tilde{\mathbf{x}}_{0,t})$\;}
         {$\Tilde{\mathbf{x}}_{0,t} \gets \Tilde{\mathbf{x}}_{0,t}$ \;}
         $z \sim\mathcal{N}(0, I)$ \;
        $x_{T_s (i-1)} \gets \sqrt{\Bar{\alpha}_{T_s(i-1)}} \hat{x}_{0} + \sqrt{1-\Bar{\alpha}_{T_s(i-1)}}z$\;
    }
    {\bfseries return} $x_0$ \;
    \end{algorithm}
\section{Evaluation}
When Robustness Evaluation \cite{RobustEvaluation} focuses on the robustness performance of the unconditional diffusion purification method\cite{DiffPure}, the evaluation for condition-guided diffusion purification is lacking. To build an effective defense mechanism, we investigate the effect of various settings to determine the most robust configuration for the conditional-guided purification method. 
Specifically, the following three factors are evaluated 1) sampling methods, 2) the number of forward steps, and 3) condition-guided methods. In addition, we also introduce and evaluate a challenging adversarial attack for condition-guided purification methods.

\subsection{Implementation Details}
\label{sec:evaluate_implementation}
 Considering ImageNet's similar results \cite{RobustEvaluation} and resource limitation, we evaluate the adversarial purification on CIFAR-10. Following the Robustness Evalulation \cite{RobustEvaluation}, we adopt Projected Gradient Descent (PGD)($\ell_{\infty},\epsilon=8/255$) \cite{PGDattack} and Expectation over Transformation (EOT) \cite{EOT} as the attack method. Some additional results are also provided in the appendix. The pre-trained WideResNet-28-10 \cite{wide_ResNet} provided by Robustbench \cite{Robustbench} is adopted as an underlying classifier. For a diffusion model, we use the pre-trained SDE model of \cite{score_based_sde}. The variances for the diffusion model are linearly increasing from $\beta_1 = 10^{-4}$ to $\beta_T = 0.02$ when $T = 1000$ \cite{ddpm}. The mediator-guided method is utilized as the condition guidance and $MSE$ is set as the distance function. We use three different sampling methods: DDPM \cite{ddpm}, DDIM~\cite{ddim}, and our random sampling.

 \begin{figure}[!t]
    \begin{center}
        \includegraphics[width=1.05\linewidth]{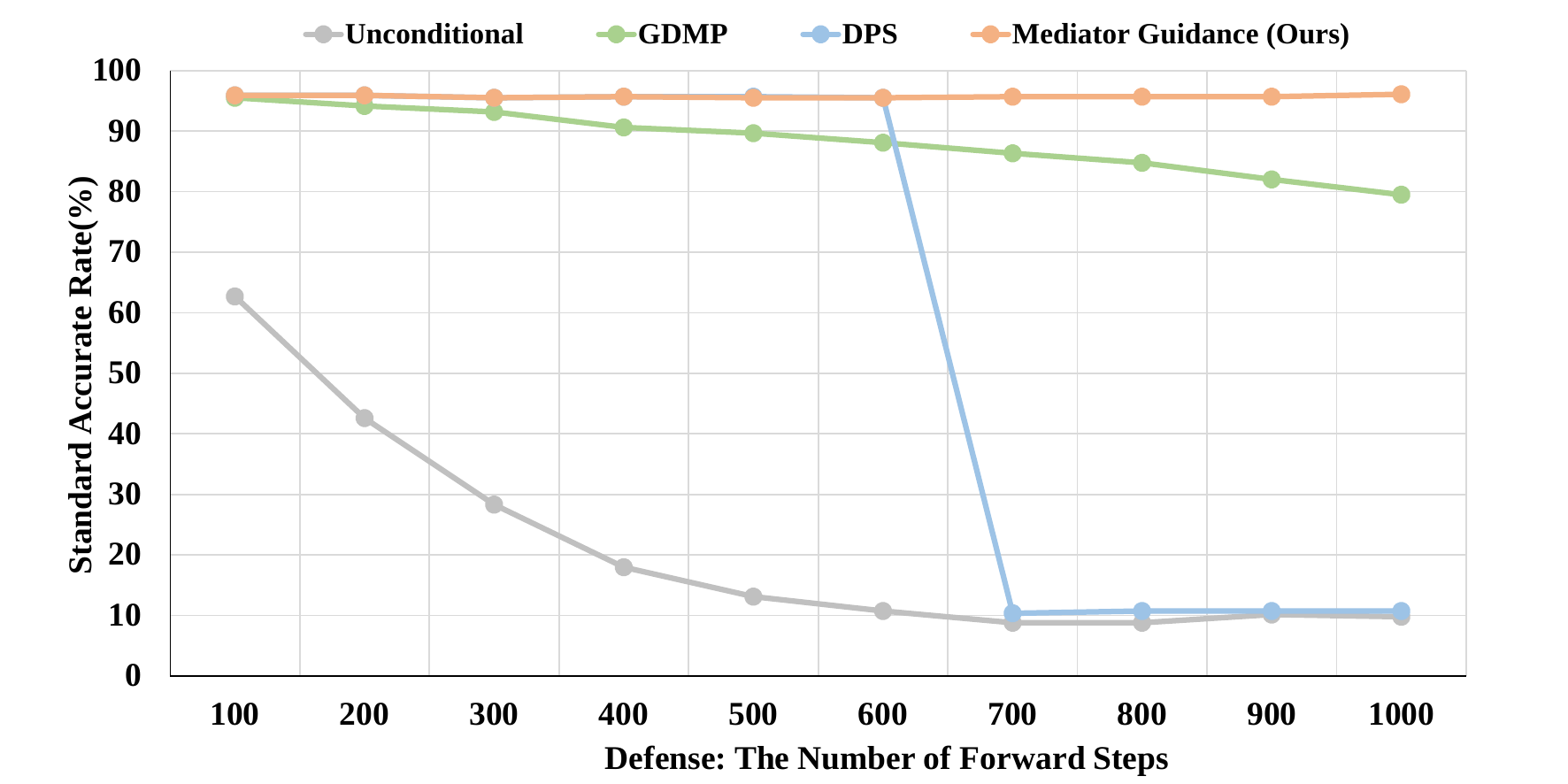}
        \vspace{-8mm}
        \caption{Standard accuracy as we change the number of defense's forward steps. 5 denoising steps for attack and 100 denoising steps for defense are used.}
        \label{fig:Guided_methods}
    \end{center}
 \vspace{-6mm}
\end{figure}

We adopt two prediction quality indexes including Standard Accuracy Rate (Standard Acc) and Robust Accuracy Rate (Robust Acc) for quantitative evaluation. Specifically, the former measures the prediction consistency, while the latter measures the defense robustness. Generally, higher values of the two mean a better purification effect.

For all experiments, we report the mean and standard deviation over five runs to measure the standard and robust accuracy. PGD uses 200 update iterations when 5 samples are used to compute EOT. Following the settings in DiffPure~\cite{DiffPure}, we use a fixed subset of 512 randomly sampled images. To calculate gradients, we use direct gradients of the entire process or surrogate process \cite{RobustEvaluation}. In each experiment, we explain the defense and attack process in detail.

\subsection{Condition Guided Method}
We evaluate several other condition guidance proposed in earlier work \cite{GDMP,MimicDiffusion,DPS} within our random sampling framework. The same number of forward steps are set for both attack and defense. Through experiments, we measure the changes in standard accuracy with the different number of forward steps in the defense, and the results are illustrated in Fig.~\ref{fig:Asynchronous_Attack}. Generally, a higher standard accuracy rate means more effective conditional guidance.

For the unconditional baseline, the standard accuracy continuously decreases as the number of forward steps increases and eventually converges to an accuracy of about $10\%$, which is fairly standard. For the 3 conditional guidance, only our proposed mediator guidance can maintain a stable accuracy and even close to the original accuracy of the classifier. The guidance of GDMP \cite{GDMP} also presents a downward trend when that of DPS \cite{DPS,MimicDiffusion} even breaks down. Thus, our proposed Mediator Guidance is a stable and effective guided method.

 \begin{figure}[!t]
    \begin{center}
        \includegraphics[width=1.05\linewidth]{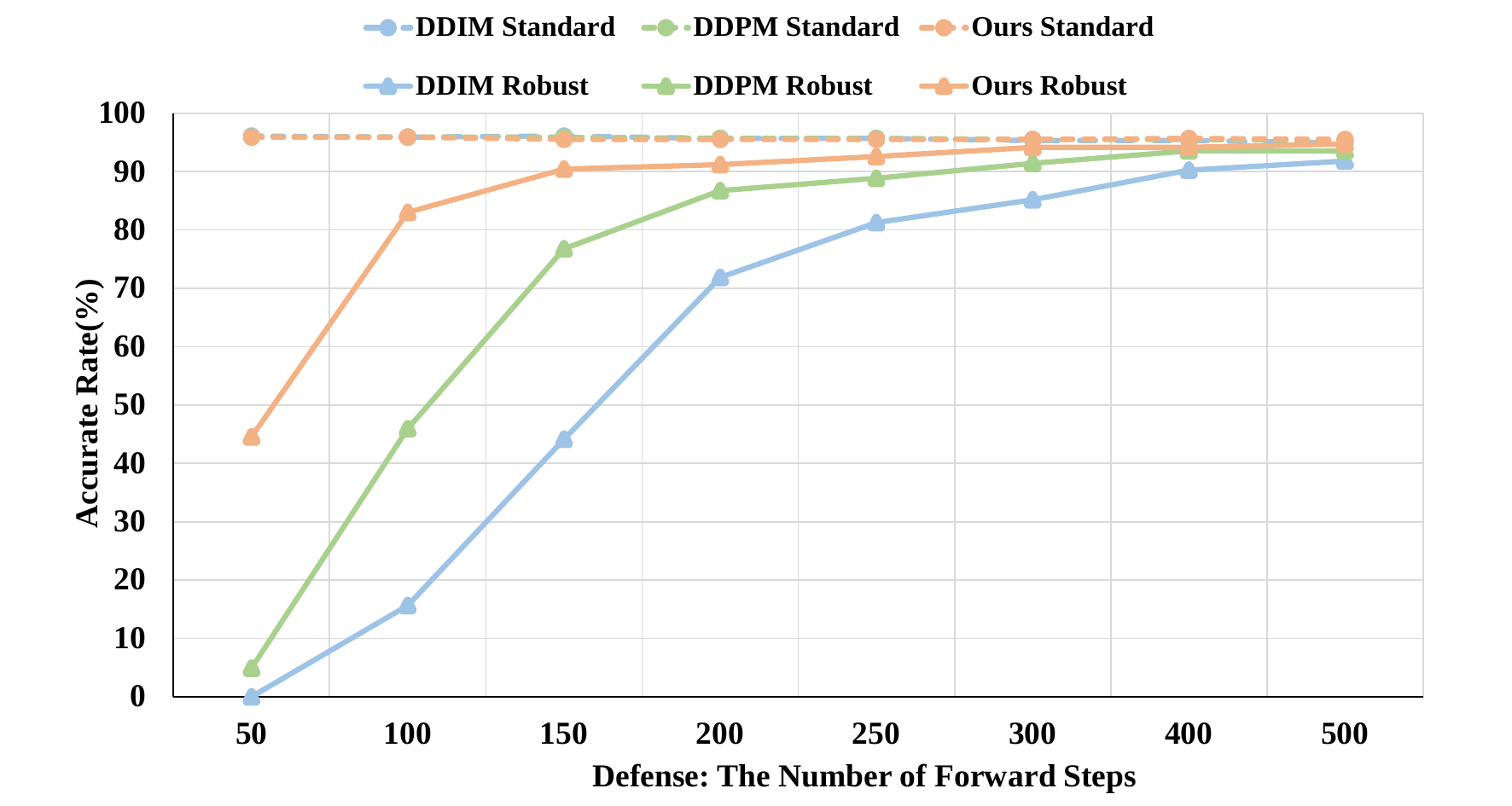}
        \vspace{-8mm}
        \caption{Standard and robust accuracy rate as we change the number of defense's forward steps ( Attack w/o Guidance). Five denoising steps for both attack and defense are used.}
        \label{fig:forward_step1}
    \end{center}
 \vspace{-7mm}
\end{figure}

 \begin{figure}[!t]
    \begin{center}
        \includegraphics[width=1.05\linewidth]{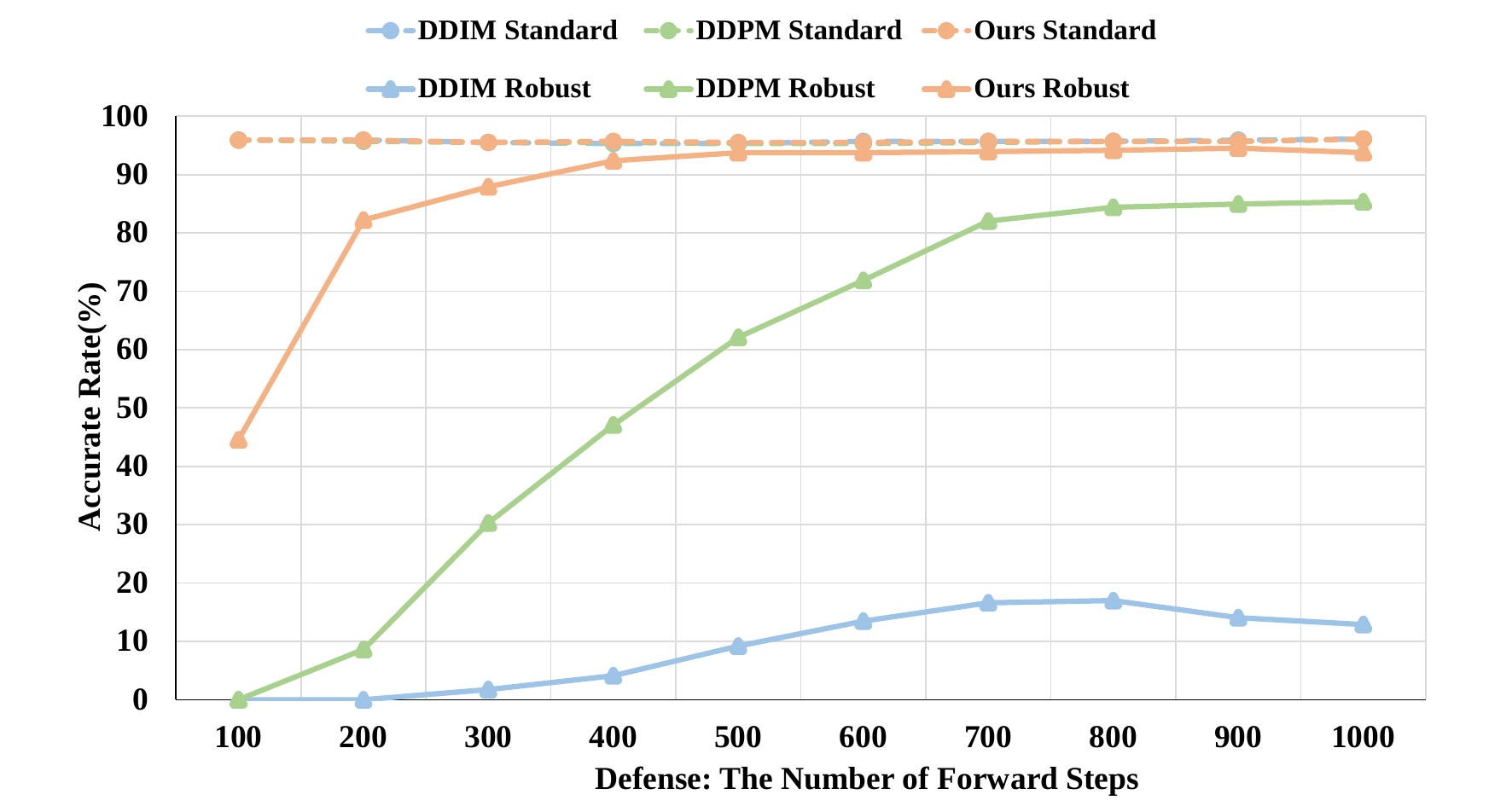}
        \vspace{-5mm}
        \caption{Standard and robust accuracy rate as we change the number of defense's forward steps ( Attack with Guidance). Five denoising steps for both attack and defense are used.}
        \label{fig:forward_step2}
    \end{center}
 \vspace{-10mm}
\end{figure}

\subsection{The Number of Forward Steps}
\label{sec: forwad_step}
We explore the effect of forward noising steps on robustness by varying the number of forward steps from 50 to 1000, resulting in changes of total variance ranging from 0.029 to 1. The same number of forward steps are used for both attack and defense, and we set five denoising steps for attack and defense for all experiments. Noting the implementation of MimicDiffusion \cite{MimicDiffusion} has a different setting than the previous work \cite{RobustEvaluation}, that is, the guided condition of the purification method is unknown to the attack method. Thus, We also conducted additional experiments and supplemented the evaluation results of unknown cases.

As shown in Fig.~\ref{fig:forward_step1} and Fig.~\ref{fig:forward_step2}, the robust accuracy continuously decreases as the number of forward steps increases since more forward steps induce more randomness, which is different from the performance of Robust Evaluation's \cite{RobustEvaluation} unconditional method. Among the three sampling methods, our random sampling achieve better robustness than commonly used DDPM sampling when the DDIM obtains the lowest robust accuracy rate, which further demonstrates the the positive correlation between randomness and robustness of sampling methods. Thus, the proposed random sampling is a more robust sampling scheme for adversarial purification. In addition, the results also show that guided conditions can improve the strength of attack methods. Under the strong attack case, the proposed random sampling method has a robustness advantage of nearly $10\%$ compared with DDPM sampling in the end, when DDIM sampling cannot effectively defend against attacks.
Meanwhile, with our proposed mediator guidance, the standard accuracy remains stable and almost achieves the original accuracy of the classifier, which intuitively shows the effect of the proposed mediation guidance method. The stable standard accurate rate curve is in sharp contrast to the continuous decline curve of unconditional guidance in Robust Evaluation \cite{RobustEvaluation}, demonstrating the superiority of the condition guidance method. 

 \begin{figure}[!t]
    \begin{center}
        \includegraphics[width=1.05\linewidth]{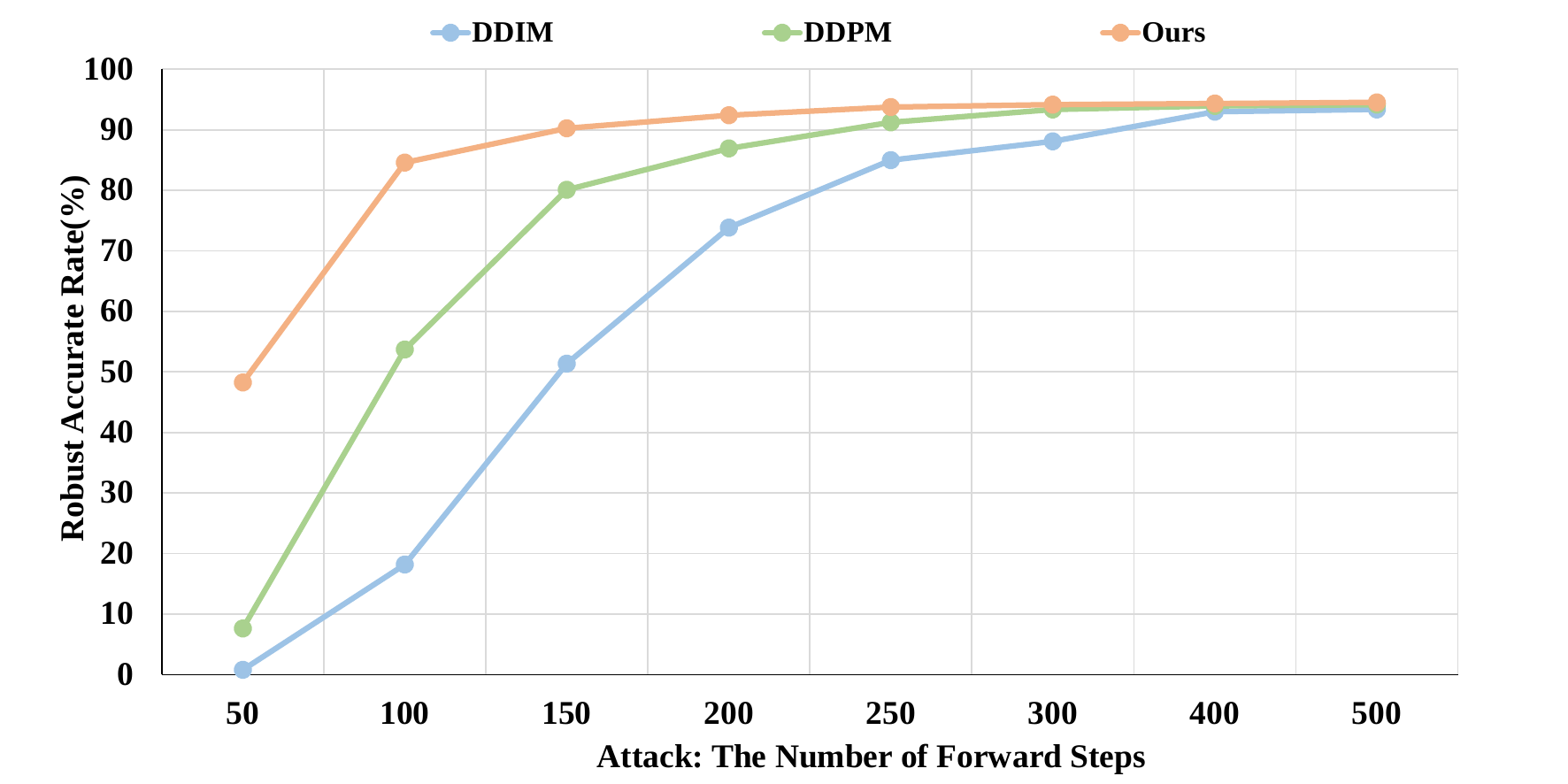}
        \vspace{-8mm}
        \caption{Robust accuracy rate as we change the number of attack's forward steps. Five denoising steps for both attack and defense are used.}
        \label{fig:Asynchronous_Attack}
    \end{center}
 \vspace{-6mm}
\end{figure}

\subsection{Asynchronous Attack}
\label{sec:asynchronous attack}
Although the Robustness Evaluation \cite{RobustEvaluation} shows that the attacker can achieve a better attack effect by using the same setup as the defender, the results may be different in the condition-guided diffusion purification methods. 
We accidentally found that the attacker can achieve a better attack effect by adopting a different number of forward steps, which we call asynchronous attacks. 
To explore the influence of the number of the attacker's forward steps, we fixed the number of the defender's forward steps at 1000, and the number of forward steps in attack ranged from 50 to 500 (more steps show the same results as 500 steps). The results are displayed in Fig.~\ref{fig:Asynchronous_Attack}.

\begin{table*}[!t]
    \centering
    \caption{Standard accuracy and robust accuracy against PGD+EOT on WideResNet-28-10.}
    \vspace{-3mm}
    \begin{tabular}{c c c c c}
        \toprule
        \textbf{Method}& \textbf{Attack Setting}  & \textbf{Denoising Steps}   &\textbf{Standard Acc} & \textbf{Robust Acc} \\
        \midrule
        DiffPure~\cite{DiffPure} & $\ell_{\infty},\epsilon=8/255$ & 100 & 91.21 $\pm$ 1.00 & 54.10 $\pm$ 3.06 \\
        GDMP~\cite{GDMP} & $\ell_{\infty},\epsilon=8/255$ & 100 & 83.01 $\pm$ 2.33 & 80.27 $\pm$ 1.82\\
        MimicDiffusion~\cite{MimicDiffusion} & $\ell_{\infty},\epsilon=8/255$ & 100 & 85.16 $\pm$ 1.29 & 83.59 $\pm$ 2.09 \\
        DiffAP (Ours) & $\ell_{\infty},\epsilon=8/255$ & 100 & \textbf{95.90} $\pm$ 1.20 & 91.02 $\pm$ 2.73 \\
        DiffAP (Ours) & $\ell_{\infty},\epsilon=8/255$ & 10 (10$\times$) & \textbf{95.90} $\pm$ 1.24 & \textbf{92.77} $\pm$ 2.05 \\
        \midrule
        DiffPure~\cite{DiffPure} & $\ell_{2},\epsilon=0.5$ & 100 & 91.21 $\pm$ 1.00 & 83.79 $\pm$ 1.08 \\
        GDMP~\cite{GDMP} & $\ell_{2},\epsilon=0.5$ & 100 & 83.01 $\pm$ 2.32 & 81.05 $\pm$ 1.51\\
        MimicDiffusion~\cite{MimicDiffusion} & $\ell_{2},\epsilon=0.5$ & 100 & 85.74 $\pm$ 1.28 & 85.93 $\pm$ 3.89 \\
        DiffAP (Ours) & $\ell_{2},\epsilon=0.5$ & 100 & \textbf{95.90} $\pm$ 1.20 & 95.12 $\pm$ 1.38 \\
        DiffAP (Ours) & $\ell_{2},\epsilon=0.5$ & 10 (10$\times$) & \textbf{95.90} $\pm$ 1.24 & \textbf{95.31} $\pm$ 1.50 \\    
        \bottomrule
    \end{tabular}
    \label{tab:normal attack}
\end{table*}

\begin{table*}[!t]
    \centering
    \caption{Standard accuracy and robust accuracy against Asynchronous PGD ($\ell_{\infty},\epsilon=8/255$) on WideResNet-28-10.}
    \vspace{-3mm}
    \begin{tabular}{c c c c c}
        \toprule
        \textbf{Method}& \textbf{Attack Setting} & \textbf{Denoising Steps}    &\textbf{Standard Acc} & \textbf{Robust Acc} \\
        \midrule
        DiffPure~\cite{DiffPure} & Asynchronous Attack & 100 & 91.21 $\pm$ 1.00 & 54.10 $\pm$ 3.06 \\
        GDMP~\cite{GDMP} & Asynchronous Attack & 100 & 83.01 $\pm$ 2.33 & 45.12 $\pm$ 4.73\\
        MimicDiffusion~\cite{MimicDiffusion} & Asynchronous Attack & 100 & 85.35 $\pm$ 1.18 & 59.57 $\pm$ 1.44 \\
        DiffAP (Ours) & Asynchronous Attack & 100 & \textbf{95.90} $\pm$ 1.20 & 82.23 $\pm$ 1.09\\
        DiffAP (Ours) & Asynchronous Attack & 10 (10$\times$) & \textbf{95.90} $\pm$ 1.20 & \textbf{83.98} $\pm$ 1.77 \\
        \bottomrule
    \end{tabular}
    \label{tab:asynchronous_attack}
\end{table*}

From the attack perspective, the results of different sampling methods demonstrate that fewer forward steps can increase the attack success rate. Surprisingly, 50 forward steps can even render the most robust defense setup based on DDPM or DDIM sampling ineffective. Therefore, the targeted asynchronous attack is a new challenge for diffusion-based adversarial purification methods.
Meanwhile, we also find that the proposed random sampling outperforms the commonly used DDPM and DDIM sampling, and gains more advantage from higher attack strength. This reveals that the random sampling method is a more robust solution for diffusion-based adversarial purification.

\section{Experiment}
Utilizing mediator-guided random sampling, we also establish a diffusion-based adversarial purification baseline called DiffAP to futher demonstrate the effectiveness of the proposed methods. The implementation follows the pipline of Algorithm~\ref{algorithm1}. And the comparison results of DiffAP with the state-of-the-art (SOTA) methods are shown below. In addition, some extended experiments are also performed to show the acceleration effect and stability.

\subsection{Experimental Settings}
\noindent\textbf{Defense Methods}. The compared diffusion-based purification methods include one unconditional method (DiffPure \cite{DiffPure}) and two conditional methods (GDMP \cite{GDMP} and MimicDiffsuion \cite{MimicDiffusion}). For a fair comparison, we adopt the evaluation setting of Sec.~\ref{sec:evaluate_implementation} for all methods. And the number of forward steps of DiffPure is set to 100 when that of other conditional methods is set to 1000.

\noindent\textbf{Attack methods}. When experimental results of previous methods \cite{DiffPure, GDMP, MimicDiffusion} have shown that the diffusion-based purification is robust enough against most attacks like AutoAttack \cite{autoattack} and backward pass differentiable approximation \cite{BPDA}, our experiment focuses on the hard case given by Robust Evaluation \cite{RobustEvaluation}, i.e., PGD+EOT Attack. The attacks involve two settings: PGD($\ell_{\infty},\epsilon=8/255$) and PGD($\ell_{2},\epsilon=0.5$) when the number of EOT samples is all set to 5. Meanwhile, We also use the stronger asynchronous attack introduced in Sec.~\ref{sec:asynchronous attack} to test the robustness of different defense methods. In a normal PGD attack, the same forward steps as the defense method and fixed 5 denoising steps are used. For the asynchronous attack, fixed 100 forward steps and 5 denoising steps are set.

\begin{table*}[!t]
    \centering
    \caption{Standard accuracy and robust accuracy at acceleration ($10\times$) and finite forward steps (100). }
    \vspace{-3mm}
    \begin{tabular}{c c c c c}
        \toprule
        \textbf{Method}& \textbf{Froward Steps} & \textbf{Denoising Steps}    &\textbf{Standard Acc} & \textbf{Robust Acc} \\
        \midrule
        DiffPure~\cite{DiffPure} & 100 & 10 & 88.67 $\pm$ 2.32 & 50.78 $\pm$ 2.04 \\
        GDMP~\cite{GDMP} & 1000 & 10 & 13.47 $\pm$ 1.51 & 14.06 $\pm$ 2.07\\
        MimicDiffusion~\cite{MimicDiffusion} & 1000 & 10 & 11.91 $\pm$ 0.95 & 10.94 $\pm$ 1.60 \\
        DiffAP (Ours) & 1000 & 10 & \textbf{95.90} $\pm$ 1.23 & \textbf{92.77} $\pm$ 2.05 \\
        \midrule
        DiffPure~\cite{DiffPure} & 100 & 100 & 91.21 $\pm$ 1.00 & 54.10 $\pm$ 3.06 \\
        GDMP~\cite{GDMP} & 100 & 100 & 95.51 $\pm$ 1.32 & 47.07 $\pm$ 2.22\\
        MimicDiffusion~\cite{MimicDiffusion} & 100 & 100 & 73.63 $\pm$ 4.17 & 43.36 $\pm$ 1.34 \\
        DiffAP (Ours) & 100 & 100 & \textbf{95.90} $\pm$ 1.24 & \textbf{83.01} $\pm$ 1.27 \\
        DiffAP (Ours) & 100 & 10 & 95.70 $\pm$ 1.29 & 82.03 $\pm$ 1.12 \\
        \bottomrule
    \end{tabular}
    \label{tab:other_comparison}
\end{table*}

\subsection{Comparison Results}
\noindent\textbf{Results for Normal Attack.}
Table.~\ref{tab:normal attack} shows the defense performance against PGD $\ell_\infty (\epsilon = 8/255)$ and $\ell_2 (\epsilon = 0.5)$ threat models, respectively. We can see that DiffAP significantly outperforms the other diffusion-based purification methods at the same denoising steps and even achieves better performance with $10\times$ sampling acceleration. Concretely, in standard accuracy rate, unconditional DiffPure performs better than the previous conditional method when our method achieves a $4.69\%$ improvement over DiffPure. This result is $10.2\%$ higher than the leading guided method MimicDiffusion and almost close to the clean accuracy of the classifier, which verifies the effectiveness of our mediator conditional guidance. Compared to MimicDiffusion on $\ell_\infty$ and $\ell_2$ PGD attacks, our DiffAP improves robust accuracy by $9.18\%$ and $9.38\%$ respectively when surpassing other methods with more than $10\%$. And it is worth noting that our method has a smaller standard deviation than other conditions methods. Thus, the results intuitively demonstrate the robustness and stability of the proposed mediator-guided random sampling. 

\noindent\textbf{Results for Asynchronous Attack.} To further distinguish the robustness of different conditional methods, the asynchronous attack is implemented in the PGD $\ell_\infty (\epsilon = 8/255)$ attack framework and the results are presented in Table~\ref{tab:asynchronous_attack}. Under the specialized strong attack, the defensive performance of conditional methods is reduced to different degrees. DiffAP has the least performance loss when other conditional methods suffer tremendous performance degradation and almost drop to the performance of unconditional DiffPure. In this case, our method achieves a huge robust accuracy advantage over other methods, which is even more $20\%$. And DiffAP still achieves the best performance with $10\times$ sampling acceleration. Different from the above normal attack, this huge performance gap is mainly caused by the difference in sampling methods, which is similar to the trend of Sec.~\ref{sec:asynchronous attack}. The results also prove that random sampling is a robust solution for adversarial purification.

\noindent\textbf{Results for Acceleration and Stability.} 
In addition, we conduct an experiment to explore the stability of the diffusion-based purification methods, mainly involving two cases of acceleration and finite forward steps \cite{RobustEvaluation}, and the results are shown in Table~\ref{tab:other_comparison}. Intuitively, our defense shows strong robustness higher than 80\% under various cases when other methods when other methods even appear to break down, which further verifies the stability of our mediator-guided random sampling. From the top results of Table~\ref{tab:other_comparison}, although our accelerated DiffAP has all achieved better results in previous results, this is not the case for other methods. Unconditional DiffPure declines in both standard accuracy rate and robust accuracy rate, when other conditional guided the method directly to collapse. We analyze that this may be because the previous conditional methods can not adapt to the varying number of denoising steps caused by, that is, they need to customize the guided conditions for each denoising step. In contrast, our mediator conditional guidance can adapt to different denoising steps without modification, which is more stable. 

Under the finite forward steps, the defensive performance of conditional methods is reduced to different degrees. As shown in the bottom of Table~\ref{tab:other_comparison}, our proposed method also outperforms all other adversarial purification methods under finite forward steps (100) with near $30\%$ robustness advantages. Other conditional methods even perform worse than the unconditional DiffPure with the same forward steps, which further the robustness and stability of the proposed guided random sampling.

\section{Conclusion}
Inspired by the DDIM sampling, we proposed an opposite sampling scheme called random sampling to enhance the robustness of diffusion-based adversarial purification. Random sampling samples from a random rather than adjacent noisy space, which brings more randomness and achieves stronger robustness against adversarial attacks. Correspondingly, a novel mediator guidance is presented to guide sampling and improve the prediction consistency, which also shows excellent stability in different Settings. To expand awareness of guided diffusion purification, we conduct a detailed evaluation with different sampling methods, which show the impressive improvement of random sampling in multiple settings. Leveraging mediator-guided random sampling, we also establish a baseline method named DiffAP, which significantly outperforms SOTA approaches in performance and defensive stability with $10\times$ sampling acceleration.

Despite the strong robustness, it is worth noting that unconditionally guided random sampling will have uncertainty, which we will explore other possibilities in further work.

{
    \small
    \bibliographystyle{ieeenat_fullname}
    \bibliography{main}
}

\clearpage
\setcounter{page}{1}

\section*{Appendix}
This document supplies more detailed derivation and experimental comparisons for comprehensive instructions.
This additional content is structured as follows: Section \ref{sec:analysis} provides a detailed analysis of the previous guided method \cite{MimicDiffusion, DPS,GDMP}. 
Sections \ref{sec:uncond_comparison} supplemented the unconditional comparison results of different sampling methods.
Sec. \ref{sec:discussion} presents the related discussion. 

\section{Analysis of Different Condition Guidance}
\label{sec:analysis}
To reveal the gradient bias introduced by the previous guidance method, we adopt $MSE$ as the distance index $d(\ast,\ast)$ for analysis. From the view of our mediator guidance, the $x_t$ could be  decomposed into the following form:
\begin{equation}
x_t = \sqrt{\bar{\alpha}_t}\Tilde{\mathbf{x}}_{0,t} + \sqrt{1-\bar{\alpha}_t} z_1, 
\end{equation}
where $\Tilde{\mathbf{x}}_{0,t} = \frac{x_{t}-\sqrt{1-\bar\alpha_t} \mathbf{\epsilon}_\theta(\mathbf{x}_{t}, t)}{\sqrt{\bar\alpha_t}}$ and $z_1$ is estimated by $\mathbf{\epsilon}_\theta(\mathbf{x}_{t}, t)$.
In GDMP\cite{GDMP}, noisy $x^{adv/clean}$ at $t$ time is obtained through one-step forward formula $q(x_t|x_0)$ \cite{ddpm} as follows:
\begin{equation}
x_t^{adv/clean} = \sqrt{\bar{\alpha}_t}x^{adv/clean} + \sqrt{1-\bar{\alpha}_t}z_2 ,\\
\end{equation}
where $z_2\sim\mathcal{N}(0,\mathbf{I})$ is a Gaussian noise different from $z_1$. Then the gradient-based guidance of GDMP \cite{GDMP} could be calculated as follows:
\begin{equation}
    \begin{aligned}
    \nabla_{x_{t}} \log p(x^{adv/clean}|x_t)=-R_{t} \nabla_{x_{t}} d\left(x_t, x_t^{adv/clean}\right) \\
    = -2R_{t}(\sqrt{\bar{\alpha}_t}(\Tilde{\mathbf{x}}_{0,t}-x^{adv/clean})+\sqrt{1-\bar{\alpha}_t}(z_1-z_2)),
    \end{aligned}
\end{equation}
where $z_1-z_2$ is the introduced gradient bias. When $R_{t}$ is a fixed constant, the time-varying part $\sqrt{1-\bar{\alpha}_t}$ of coefficient gradually increases with the increase of time $t$, the guiding error naturally increases, which is also consistent with the experimental results of main paper Fig.~3.

In DPS \cite{DPS, MimicDiffusion},  the gradient-based guidance is employed to guide the update of $x_{t-1}$ as follows: 
\begin{equation}
   x_{t-1} \gets x_{t-1} - \nabla_{x_{t}} \log p(x^{adv/clean}|x_t),
\end{equation}
which brings a bias when using the gradient at $t$ time to guide the update at $t-1$ time and the gradient guidance is calculated as follows:
\begin{equation}
\begin{aligned}
\nabla_{x_{t}} \log p(x^{adv/clean}|x_t) = -R_{t} \nabla_{x_{t}} d\left(\Tilde{\mathbf{x}}_{0,t}, x^{adv/clean}\right) \\
= \frac{-2R_{t}}{\sqrt{\bar\alpha_t}}(\Tilde{\mathbf{x}}_{0,t}-x^{adv/clean}).
\end{aligned}
\end{equation}
Considering the reasonable guidance is supposed to be as follows:
\begin{equation}
   x_{t-1} \gets x_{t-1} - \nabla_{x_{t-1}} \log p(x^{adv/clean}|x_{t-1}).
\end{equation}
Thus, the introduced gradient bias is $(\frac{{2R_{t-1}}\Tilde{\mathbf{x}}_{0,t-1}}{{\sqrt{\bar\alpha_{t-1}}}}-\frac{2R_{t}\Tilde{\mathbf{x}}_{0,t}}{\sqrt{\bar\alpha_t}})+(\frac{2R_{t}}{\sqrt{\bar\alpha_t}}-\frac{2R_{t-1}}{\sqrt{\bar\alpha_{t-1}}})x^{adv/clean}$. Possibly because the larger time $t$ is, the closer the molecule ${\bar\alpha_{t-1}}$ is to 0 and the greater the error in the estimation of $\Tilde{\mathbf{x}}_{0,t}$, the DPS guidance crashes when the time $t$ is greater than a certain value.

 \begin{figure}[!t]
    \begin{center}
        \includegraphics[width=1.05\linewidth]{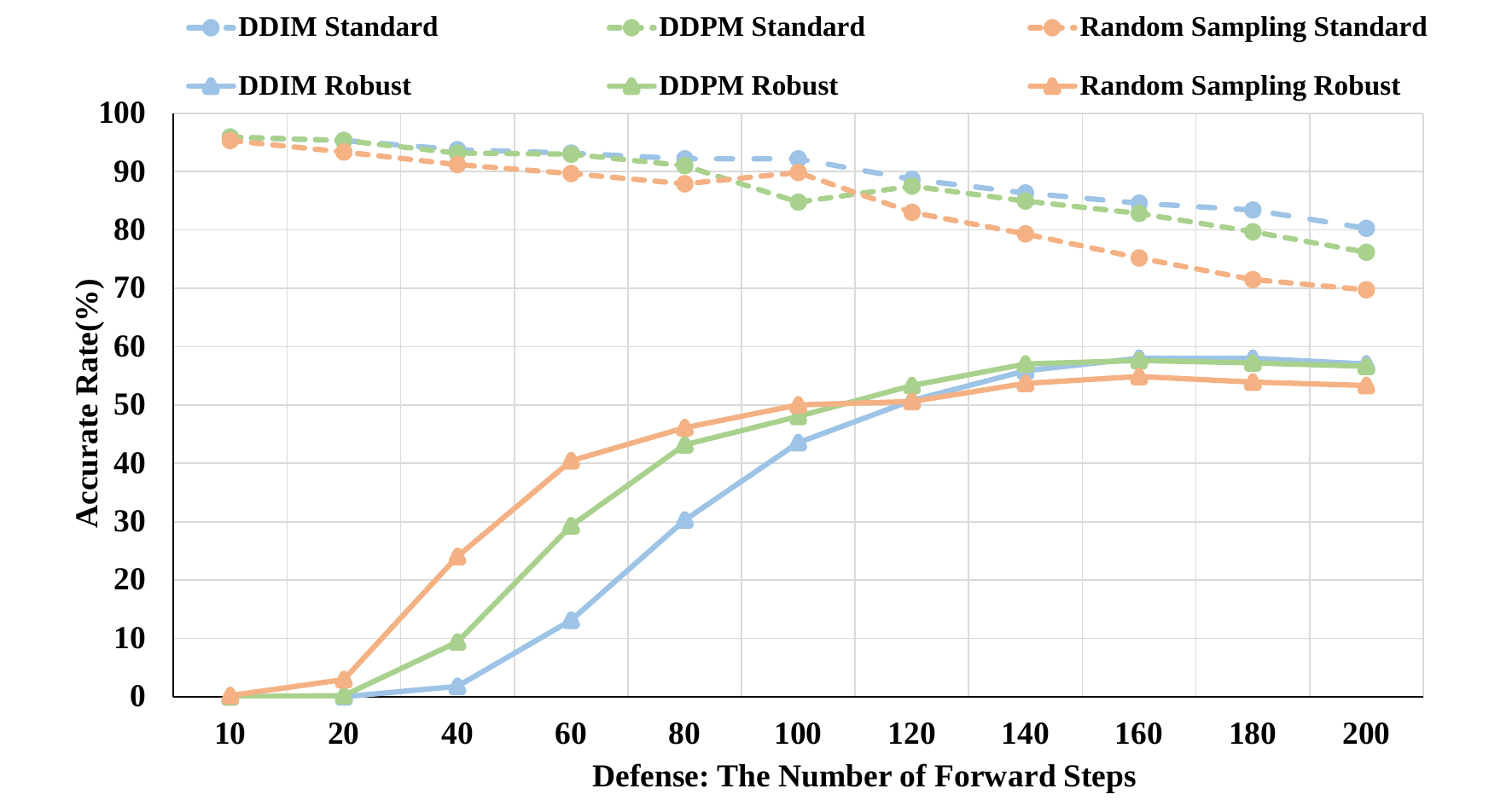}
        \vspace{-8mm}
        \caption{Standard and robust accuracy rate of unconditional cases as we change the number of defense's forward steps.}
        \label{fig:forward_step_uncond}
    \end{center}
 \vspace{-7mm}
\end{figure}

\section{Unconditional Sampling Comparison}
\label{sec:uncond_comparison}
To comprehensively show the difference between different sampling methods, we supplemented the results without condition guidance as Fig.~\ref{fig:forward_step_uncond}. The number of forward steps varies from 10 to 200, which results in changes of total variance ranging from 0.0014 to 0.2364. The same number of forward steps are used for both attack and defense, and we set five denoising steps for attack and defense for all experiments. Other experimental settings are the same as those in the main paper.

Similar to the results of the robustness evaluation \cite{RobustEvaluation}, the standard accuracy of all sampling methods showed a downward trend. The magnitude of the decline is proportional to the randomness of the sampling method, and DDIM exhibits the highest standard accuracy, which is yet much lower than the performance of conditional guidance. The robustness accuracy of all sampling methods showed a trend of first increasing and then decreasing, with the highest accuracy at 160 forward steps. From the results, the robust accuracy of random sampling increases faster with the number of forward steps, but the final robust accuracy is slightly lower. In contrast, the robust accuracy of DDIM increases slowly with the number of forward steps, but the final robust accuracy is the highest.
We think the standard accuracy of all methods is relatively high when the number of forward steps is small, and the robustness accuracy mainly depends on the robustness of the sampling method. When the number of forward steps is large, the decline in standard accuracy will limit the upper limit of robust accuracy, resulting in stable methods eventually achieving high robust accuracy, which is why this phenomenon occurs. It is worth noting that the best robustness performance of unconditional method is also far from the condition guided methods of main paper.

\section{Discussion}
\label{sec:discussion}
As demonstrated in \cite{Robustbench}, finding the worst case is important for defense methods. Thus, asynchronous attacks are introduced to challenge existing diffusion-based purification methods. Until now, diffusion-based adversarial purification methods have had no negative social impact. Our proposed random sampling method does not present any negative foreseeable societal consequence, either.

\end{document}